\documentclass[runningheads]{llncs}

\usepackage{eccv}

\usepackage{eccvabbrv}

\usepackage{graphicx}
\usepackage{booktabs}

\usepackage[accsupp]{axessibility}  % Improves PDF readability for those with disabilities.

\usepackage{hyperref}

\usepackage{orcidlink}
\usepackage[linesnumbered,ruled,vlined]{algorithm2e}

\begin{document}
% ---------------------------------------------------------------
% TODO REVIEW: Replace with your title
\title{Enhancing Thermal MOT: A Novel Box Association Method Leveraging Thermal Identity and Motion Similarity} 

% TODO REVIEW: If the paper title is too long for the running head, you can set
% an abbreviated paper title here. If not, comment out.
%\titlerunning{Abbreviated paper title}

% TODO FINAL: Replace with your author list. 
% Include the authors' OCRID for the camera-ready version, if at all possible.
\author{Wassim Ali El Ahmar\inst{1} \and
Dhanvin Kolhatkar\inst{1} \and
Farzan Nowruzi\inst{1} \and
Robert Laganiere\inst{1}}

% TODO FINAL: Replace with an abbreviated list of authors.
\authorrunning{W.~El Ahmar et al.}
% First names are abbreviated in the running head.
% If there are more than two authors, 'et al.' is used.

% TODO FINAL: Replace with your institution list.
\institute{University of Ottawa, Ottawa ON K1N 6N5, Canada}

\maketitle

\begin{abstract}
Multiple Object Tracking (MOT) in thermal imaging presents unique challenges due to the lack of visual features and the complexity of motion patterns. This paper introduces an innovative approach to improve MOT in the thermal domain by developing a novel box association method that utilizes both thermal object identity and motion similarity. Our method merges thermal feature sparsity and dynamic object tracking, enabling more accurate and robust MOT performance. Additionally, we present a new dataset comprised of a large-scale collection of thermal and RGB images captured in diverse urban environments, serving as both a benchmark for our method and a new resource for thermal imaging. We conduct extensive experiments to demonstrate the superiority of our approach over existing methods, showing significant improvements in tracking accuracy and robustness under various conditions. Our findings suggest that incorporating thermal identity with motion data enhances MOT performance. The newly collected dataset and source code is available at \href{https://github.com/wassimea/thermalMOT}{https://github.com/wassimea/thermalMOT}
\end{abstract}
    
\section{Introduction}
\label{sec:intro}

Thermal cameras have proven to be robust perception sensors that operate reliably under different weather and lighting conditions for various tasks in computer vision~\cite{Lee2015, Lahouli2018, Nowruzi2019, Kutuk2022, Shin2023, Rivadeneira2023, Dai2020, Broyles2022, Brenner2023}. This characteristic of thermal cameras allows vision systems utilizing them to take advantage of the unique thermal patterns of objects for more robust and reliable performance. Convolutional neural networks (CNNs) have been used to great effect for a variety of computer vision tasks for different spectrums (RGB, thermal, depth, hyperspectral, etc.) These tasks range from image classification~\cite{He2015, Xie2016, Tan2019, Liu2022} to object detection~\cite{Law2018, Duan2019, Wang2020} and multiple object tracking~\cite{Wojke2017, Wang2019, Zhang2020, bytetrack}.

%-------------------------------------------------------------------------
\subsection{Multiple Object Trackers}

Multiple Object Tracking (MOT) is the task of detecting individual objects in a video and tracking them over consecutive frames with a unique identifier. The performance of a network is measured in terms of how each object is tracked over multiple frames with a consistent ID.

Solutions to the multiple object tracking (MOT) task can be divided into two main categories: one-stage~\cite{Feichtenhofer2017, Zhou2020a, Sun2020a, Lu2020} and two-stage~\cite{Bewley2016, Wojke2017, Wang2019, Zhang2020, bytetrack, Cao2023}. The former type uses an end-to-end pipeline that tracks directly from the network's inputs while the latter separates the task into (1) the detection of objects in the scene and (2) the tracking of these detections in subsequent frames. Two-stage approaches have proven to be more versatile and accurate than one-stage trackers~\cite{Zhang2020}. %While these tracking methods can be used directly on thermal images, this does not take advantage of characteristics unique to the thermal spectrum: that of the thermal signature that results from a material's specific physical characteristics (specifically its diffusion of heat).

Most two-stage MOT solutions utilize motion association as the main criteria when conducting box association: a Kalman filter~\cite{Kalman1960} is used to predict the locations of objects in the next frame; Intersection-over-Union (IoU) is then calculated between the detected bounding boxes and the Kalman-filter predicted boxes to match the boxes across frames. One main advantage of motion association is that the algorithm can be utilized in systems using any type of sensor (visible and non-visible): as long as it is possible to predict bounding boxes, it is possible to conduct tracking, regardless of the sensor modality.

However, strictly relying on motion association does have an important drawback: not utilizing the unique characteristics of any one sensor. For example, pixel proximity and distance information is valuable for conducting box association with LiDAR or 3D camera data. In our work, we show that utilizing the thermal identity of objects in the two-stage trackers' tracklet association step can lead to more accurate box association. Instead of strictly relying on motion association, we devise a box association algorithm that utilizes both the motion information and the thermal identity of objects.  

The MOT challenges~\cite{LealTaixe2015, Milan2016, Voigtlaender2019, Dendorfer2019, Dendorfer2020, Dave2020, Pedersen2020} provide very popular benchmarks for MOT with RGB images. Their datasets offer a wide variety of scenes, including many busy pedestrian sequences. However there is a lack of large public datasets for the MOT task with thermal images, severely limiting research on this task.

\subsection{Contributions}
The main contributions of this paper are:
\begin{itemize}
	\item The development, annotation, and forthcoming public release of a unique dataset that, to the best of our knowledge, stands as a significant contribution comparable in size and urban environment settings to the MOT17 benchmark. Uniquely, our dataset integrates matching RGB and thermal images across five different pedestrian crossing locations, offering a comprehensive resource for both detection and multi-object tracking tasks.
	\item We introduce a novel box association method for use with two-stage MOT models running in the thermal spectrum that utilizes the unique characteristics of thermal data and combines them with motion data for robust MOT in the thermal spectrum. Although this novel box association method focuses on unique characteristics of thermal imagery, we believe our work would encourage the research and development of algorithms that leverage the unique attributes of any sensor when conducting MOT. Thereby, our approach is generalizable to other sensor modalities as well.
	\item We provide an initial benchmark of the performance of two state-of-the-art two-stage MOT models on both modalities of our dataset (RGB and thermal).
\end{itemize}

\section{Literature Review}
\label{sec:litreview}

Our work builds upon the existing literature of object detection, two-stage multiple object tracking, the use of thermal sensors for computer vision tasks, and existing MOT datasets with thermal images. %This section presents this previous research.

%-------------------------------------------------------------------------
\subsection{Object Detection}

The task of object detection consists of the detection in a scene of each object that belongs to a set list of categories. The approaches can be divided into two main categories depending on whether the detection pipeline generates object proposals to be refined by a second stage (two-stage object detection)~\cite{Girshick2014, Girshick2015, Ren2015, Cai2017, He2017, Duan2019} or not (one-stage)~\cite{Redmon2015, Liu2015, Lin2017, Law2018}. Models in the latter category include SSD~\cite{Liu2015} and YOLO~\cite{Redmon2015}, and generally offer a simpler and faster architecture, while still achieving results competitive with two-stage models.

The Task-aligned One-stage Object Detection (TOOD) network, designed by Feng et al.~\cite{tood}, iterates upon preceding one-stage object detection networks. Its architecture includes the Task-aligned Head (T-Head) which improves feature sharing for the classification and localization sub-tasks, as opposed to using separate network heads for each one. The Task-aligned Predictor (TAP) part of the T-head then improves the alignment of the classification and localization predictions to better combine them. The training of TOOD is also modified through Task Alignment Learning (TAL) which improves default anchor proposals.

%-------------------------------------------------------------------------
\subsection{Two-Stage Multiple Object Trackers}

Two-stage multi-object trackers can be divided into two sub-tasks, one for each stage in the MOT solution: (1) per-frame object detection and (2) tracking over a sequence. This tracking-by-detection approach enables the easy and direct use of existing state-of-the-art object detection networks as a high-accuracy first stage and places the design focus on the tracking stage~\cite{Bewley2016, Wojke2017, Wang2019, Zhang2020, bytetrack, Cao2023}. This also allows the training of the first stage on object detection datasets that do not necessarily have tracking annotations, decoupled from the MOT task. The main downside of this is in the limitations of the second stage to recover from mistakes in the detection stage, whether false positives or false negatives.

The Simple Online and Realtime Tracking (SORT) approach was introduced by Bewley et al.~\cite{Bewley2016} and combines any of a variety of CNN object detectors with a straightforward motion estimation tracking approach. After detection in a frame, a Kalman filter~\cite{Kalman1960} is used to approximate the velocity and the future location of each tracklet. For a new frame, the IoU between the approximated location of the tracklets and the detector's predictions is used to assign either an existing tracklet identification or a new ID altogether. Wojke et al.'s DeepSORT~\cite{Wojke2017} improved upon the SORT approach by introducing a CNN-based motion and appearance association metric that strongly improves performance.

Association methods can rely strongly on a detected object's confidence score to decide on which boxes to use~\cite{Wang2019, Zhang2020, Lu2020, Sun2020a}. Going against this approach, Zhang et al.~\cite{bytetrack} introduced the Bytetrack tracker using their BYTE association method which uses detections whether they have high or low confidence. Specifically, low-confidence detections are not discarded but are instead given lower matching priority with existing tracklets.

OCSort, designed by Cao et al.~\cite{Cao2023}, improves directly on shortcomings of the SORT approach in tracking occluded objects, especially when their motion is non-linear. To combat the issue of error accumulation when tracking an object that has been temporarily lost, the authors propose the Observation-centric Re-Update (ORU) strategy. ORU leverages observations from a virtual trajectory to recalibrate the parameters of the Kalman filter. Furthermore, the authors introduce an Observation-Centric Momentum (OCM) term within the association cost function, prioritizing the use of observations over estimations to improve the accuracy of motion direction estimation. An Observation-Centric Recovery (OCR) technique is also introduced, which aids in recovering temporarily lost object tracks, such as those occluded or momentarily stationary. Together, these methodological innovations constitute the OC-SORT algorithm, which is developed as an enhancement to ByteTrack~\cite{bytetrack}.

%-------------------------------------------------------------------------
\subsection{Thermal Sensors}
Thermal sensors are being used to great effect for a variety of computer vision tasks either on their own~\cite{Lee2015, Lahouli2018, Nowruzi2019, Kristo2020, us_2022, Kutuk2022, Shin2023, Rivadeneira2023} or through the use of sensor fusion with RGB cameras ~\cite{Dai2020, Broyles2022, Brenner2023, Ahmar2023a}. 

Lee et al.~\cite{Lee2015} and Lahmyed et al.~\cite{Lahmyed2018} both use an RGB camera and a thermal camera for the detection of pedestrians. In the former, motion is detected independently in each sensor, and is then used to predict pedestrian location in both sensors.

Lahouli et al.~\cite{Lahouli2018} devise a low computational cost method of tracking pedestrians using compressed thermal images taken by Unmanned Aerial Vehicles (UAVs). The suggested approach follows a tracking-by-detection framework where Regions of Interest (ROIs) are proposed and refined using saliency maps and contrast enhancement techniques. These are then tracked over consecutive images using the MPEG compression algorithm's motion vectors.

Nowruzi et al.~\cite{Nowruzi2019} propose a method to detect the number of passengers in a vehicle. A low-cost CNN is used to detect individuals while meeting the requirements for use in an embedded system. Thermal images are used for their privacy-preserving quality, as it is much harder to identify individuals from the features present in a thermal image than an RGB image.

Ahmar et al.~\cite{us_2022, Ahmar2023a} collected a dataset of matching RGB and thermal frames with object detection and MOT annotations. The authors used this dataset to compare detection and tracking performance from RGB and thermal data. In addition, they studied the use of multi-modal sensor fusion from the two modalities and proposed a new fusion method which noticeably increases object detection performance over existing fusion approaches.

%-------------------------------------------------------------------------
\subsection{Datasets}
Most existing tracking datasets that include thermal images are annotated for single-object tracking~\cite{Liu2020, Liu2020a, Li2022}. For the task of MOT, the City-Scene dataset~\cite{us_2022, Ahmar2023a} contains 15 sequences for a total of 1 997 annotated frames for both a FLIR thermal camera and an RGB camera. However, none of the existing thermal MOT datasets offer a sufficient volume of data for multi-object tracking for pedestrians. With limited sequences and annotated frames, these datasets are inadequate for training and evaluating MOT algorithms in complex real-world scenarios. Recognizing this limitation, we undertook the collection and annotation of a new, large-scale dataset containing both RGB and corresponding thermal data. This initiative addresses the need for a more extensive and diverse dataset collected in real-world scenarios, allowing for the development and assessment of MOT algorithms in challenging, real-world conditions in both the thermal and color spectrums.

%PTB-TIR~\cite{Liu2020} collects 60 thermal-only sequences from various cameras for a total of 30128 frames, LSOTB-TIR~\cite{Liu2020a}. LasHer~\cite{Li2022}

\section{Data Collection}
\label{sec:dataCollection}

We utilize a FLIR ADK thermal sensor and a JAI GO-5100C RGB sensor for the collection of the new dataset. The FLIR ADK thermal sensor specifications include a 75° horizontal field of view, operates in the 8-14 microns (LWIR) spectral band, has a thermal sensitivity of less than 50 mK, consumes an average of 4W of power, and offers an image resolution of 640x512. The JAI GO-5100C RGB sensor, it features a global shutter, consumes 4.35W of power, and can achieve frame rates of up to 74 frames per second. A plastic enclosure was built to fix both sensors next to each other, and was mounted on a tripod for data collection. 

\begin{figure*}
	\centering
	\begin{subfigure}{0.315\textwidth}
		\includegraphics[width=\linewidth]{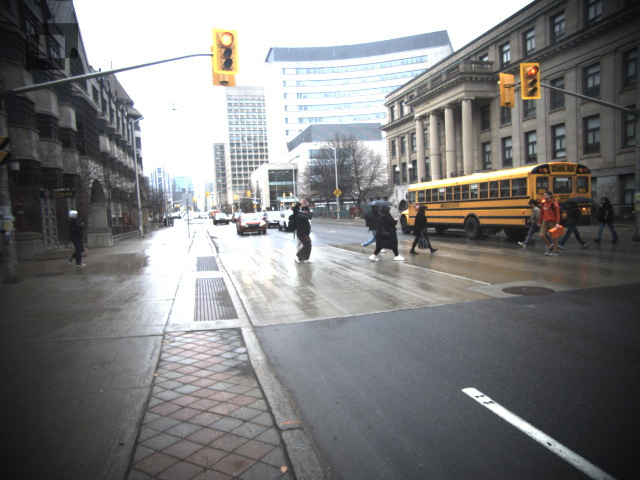}
	\end{subfigure}
	\hfill
	\begin{subfigure}{0.315\textwidth}
		\includegraphics[width=\linewidth]{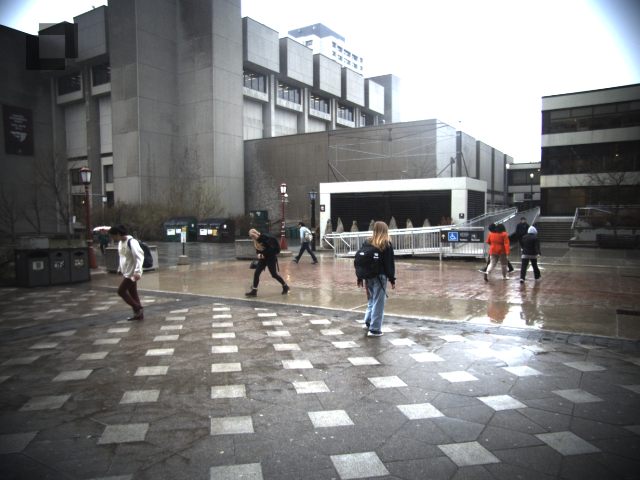}
	\end{subfigure}
	\hfill
	\begin{subfigure}{0.315\textwidth}
		\includegraphics[width=\linewidth]{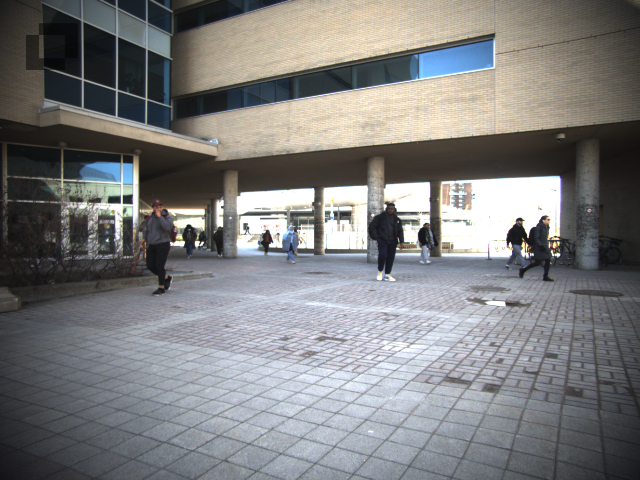}
	\end{subfigure}

	\begin{subfigure}{0.315\textwidth}
		\includegraphics[width=\linewidth]{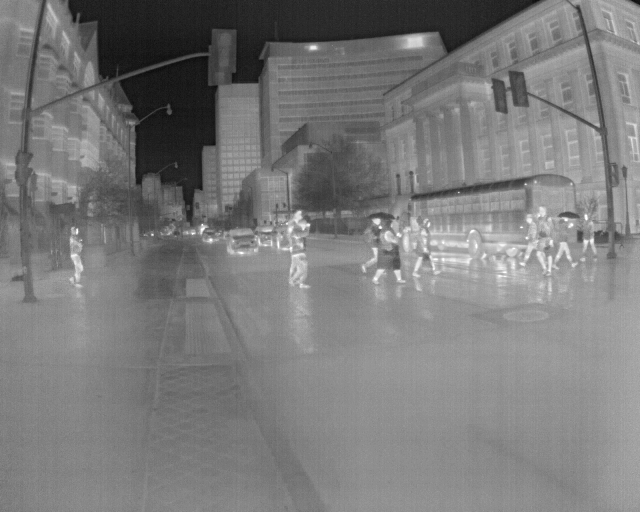}
	\end{subfigure}
	\hfill
	\begin{subfigure}{0.315\textwidth}
		\includegraphics[width=\linewidth]{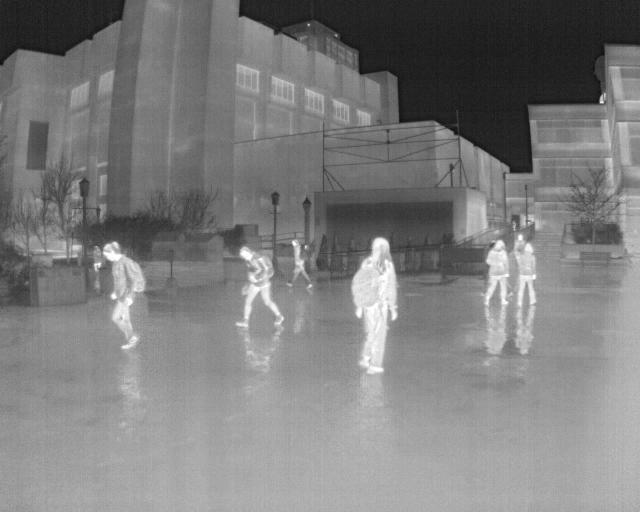}
	\end{subfigure}
	\hfill
	\begin{subfigure}{0.315\textwidth}
		\includegraphics[width=\linewidth]{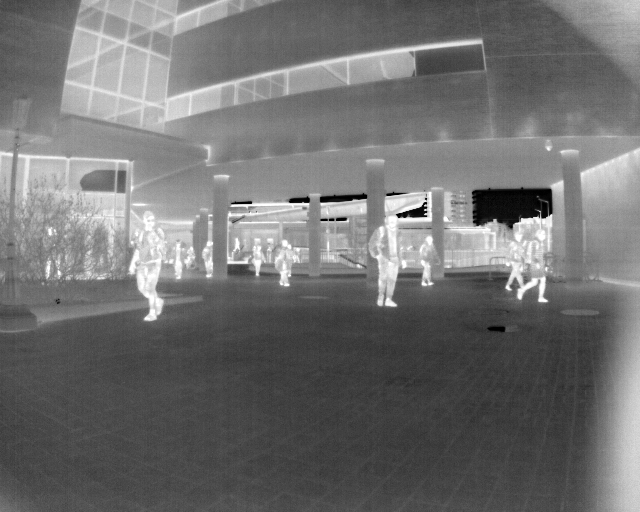}
	\end{subfigure}
	
	%\begin{subfigure}{0.48\textwidth}
	%	\includegraphics[width=0.4\linewidth]{Figures/seq4_rgb}
	%\end{subfigure}
	%\hfill
	%\begin{subfigure}{0.48\textwidth}
	%	\includegraphics[width=0.4\linewidth]{Figures/seq4_thermal}
	%\end{subfigure}

	\caption{Sample RGB and corresponding thermal images from the RGB-Thermal MOT dataset.}
	\label{figure_tmot_dataset}
\end{figure*}

The dataset of 30 sequences (9000 frames per modality total) was collected at 5 different intersections around an urban campus in public spaces. RGB and thermal samples from the dataset are shown in Figure~\ref{figure_tmot_dataset}. The dataset was then annotated for multiple object detection and tracking. We refer to this dataset as the RGB-Thermal MOT dataset. 

To the best of our knowledge, this is the world's first large-scale dataset of RGB and corresponding thermal images annotated for MOT. We believe that this dataset will prove to be a valuable resource for research and development of both thermal and RGB MOT research. Additional statistics related to the dataset are given in Table~\ref{dataset_statistics_table}.

In the collection of our RGB-Thermal pedestrian dataset, we diligently followed local regulations and ethical guidelines to ensure the respectful and responsible use of data. We sought advice from local authorities to align with privacy and data protection standards, emphasizing the responsible use of urban imagery. The dataset, aimed for pedestrian detection and multi-object tracking research, was collected under conditions that respect public space and individual privacy. This process reflects our commitment to ethical research practices, contributing to the field's advancement while upholding high ethical standards.

%\begin{figure}
%	\begin{center}
%		\includegraphics[width=0.3\textwidth]{Figures/tripod.jpg}
%		\caption{Capturing system}
%		\label{capturing_system}
%	\end{center}
%\end{figure}

%\begin{figure}
%	\begin{center}
%		\includegraphics[width=0.3\textwidth]{Figures/mount.jpg}
%		\caption{Sensor plastic enclosing}
%		\label{mount}
%	\end{center}
%\end{figure}

\begin{table*}[t!]
	\centering
	\begin{tabular}{|c|c|c|}
		\hline
		& \textbf{Thermal}   & \textbf{RGB}       \\ \hline
		\textbf{Total number of annotated sequences}             & 30 (1 minute each) & 30 (1 minute each) \\ \hline
		\textbf{Capture frame rate}                              & 5                 & 5                 \\ \hline
		\textbf{Total frames per sequence}                       & 300                & 300                \\ \hline
		\textbf{Total number of annotations}                     & 58,590             & 50,400             \\ \hline
		\textbf{Average annotations per image}                   & 6.51               & 5.6                \\ \hline
		\textbf{Sequence Train/Test split}               & Train: 24          & Train: 24          \\
		& Test: 6            & Test: 6            \\ \hline
		\textbf{Total unique tracks Train/Test} & Train: 313         & Train: 284         \\
		& Test: 126          & Test: 116          \\ \hline
	\end{tabular}
	\caption{RGB-Thermal MOT dataset statistics.}
	\label{dataset_statistics_table}
\end{table*}

\section{Experiments}
\label{sec:experiments}
Our study focuses specifically on the enhancement of the box association step for Multiple Object Tracking (MOT) in thermal imagery. We leverage the capabilities of the two most advanced two-stage trackers, (ByteTrack~\cite{bytetrack} and OCSORT~\cite{Cao2023}), as the basis for our experiments. This focused approach allows us to isolate and evaluate the impact of our novel thermal box association method distinctly.

Our work is aimed at advancing the precision of box association in thermal MOT scenarios, a critical area that benefits from the integration of thermal characteristics into tracking algorithms. By maintaining a fixed detection framework, our experiments directly attribute observed enhancements in tracking accuracy to our proposed method. This approach ensures that the contribution of incorporating thermal information into the box association step is rigorously assessed, highlighting its significance in the advancement of MOT technologies.

In our work, we fine-tune the TOOD~\cite{tood} detector with a Resnet50~\cite{resnet50} backbone on images from the newly collected RGB-Thermal MOT dataset. We use the OpenMMlab mmdetection framework~\cite{mmdetection} for training the object detector. The outcome of this training is two models:
\begin{itemize}
	\item TOOD model initialized with COCO weights and fine-tuned on the RGB images of the RGB-Thermal MOT dataset.
	\item TOOD model initialized with COCO weights and fine-tuned on the thermal images of the RGB-Thermal MOT dataset.
\end{itemize}

\subsection{Benchmarking SOTA MOT models on the RGB-Thermal MOT dataset}
We use the OpenMMlab mmtracking framework~\cite{mmtracking} to benchmark two SOTA MOT models (ByteTrack~\cite{bytetrack} and OCSORT~\cite{Cao2023}), on both the RGB and thermal sequences of the RGB-Thermal MOT dataset. The corresponding RGB/thermal TOOD detectors described are used as backbones for the trackers. 

We optimize the hyperparameters of both ByteTrack and OCSORT to maximize their MOT metrics (specifically MOTA and IDF1) when using their standard implementations. This step is crucial to ensure a fair comparison of our innovative box association approach against the highest achievable performance of each of those two trackers.

Furthermore, we conduct a performance evaluation of ByteTrack and OCSORT on the RGB sequences within the RGB-Thermal MOT dataset. This assessment serves as a valuable reference for future users of this recently acquired dataset.

\subsection{Development of New MOT Box Association Method Utilizing Thermal Information}

For enhancing Multiple Object Tracking (MOT) capabilities in thermal imagery, we introduce a novel method that leverages motion and thermal identity characteristics of detected objects to enhance the quality of MOT box association in thermal imagery. This approach aims to capitalize on the unique, yet sparse, thermal signatures captured in thermal imaging, a dimension that standard MOT models typically overlook.

The core of our proposed algorithm is the integration of thermal and motion data to establish a robust tracking framework, as outlined below:

\paragraph{Thermal and Motion Bounding Boxes:}
Let \( T = \{t_1, t_2, ..., t_m\} \) and \( D = \{d_1, d_2, ..., d_n\} \) denote the sets of tracking boxes predicted using a Kalman filter, and detection bounding boxes, respectively, within a given thermal image, denoted as \(I_{thermal}\).

\paragraph{Initialization:}
The conversion of bounding box coordinates into NumPy arrays is a fundamental step for efficient manipulation and computation within our algorithm:
\begin{equation}
T, D \leftarrow \text{numpy}(T, D)
\end{equation}
This step lays the groundwork for the algorithm's operations.

\paragraph{Similarity Matrix Construction:}
We define \( S_{thermal} \), an \( m \times n \) matrix, to quantify the thermal similarity between each pair of \( t_i \) and \( d_j \). Initially,
\begin{equation}
S_{thermal} \leftarrow \text{Zeroes}(m, n)
\end{equation}
This initialization is a preparatory step for accumulating similarity scores, rather than indicating an absence of similarity.

\paragraph{Pairwise Histogram Comparison:}
For each \( t_i \) in \( T \) and \( d_j \) in \( D \), their respective Regions of Interest (ROIs) are extracted from \(I_{thermal}\), denoted as \( ROI_{t_i} \) and \( ROI_{d_j} \). The histograms \( H_{t_i} \) and \( H_{d_j} \) are then calculated, normalized to \( \hat{H}_{t_i} \) and \( \hat{H}_{d_j} \), using appropriate bin sizes and ranges to capture the thermal characteristics. The normalization process ensures histograms are on a uniform scale, essential for accurate comparison. The similarity \( s_{ij} \) is computed using the Bhattacharyya coefficient, a robust measure for histogram comparison, to populate \( S_{thermal} \) with meaningful values.

\paragraph{Integration with Motion Similarity:}
The motion-based similarity matrix, \( S_{motion} \), obtained through standard MOT methodologies, is integrated with \( S_{thermal} \) to form the comprehensive similarity matrix, \( S_{comp} \):
\begin{equation}
S_{comp} = \alpha \cdot S_{motion} + (1 - \alpha) \cdot S_{thermal}
\end{equation}
Here, \( \alpha \) represents a carefully selected weighting factor that balances the contributions of motion and thermal similarities, optimized through experimental validation to ensure effective tracking performance. For ByteTrack, the optimal value of \( \alpha \) is proven to be $0.3$, and that for OCSort is proven to be $0.8$ (Figure \ref{figure_bytetrack_ocsort_alpha}). The value of \( \alpha \) is selected as to ensure the best trade-off between MOTA and IDF1.

The pseudo-code of the function is given in Algorithm~\ref{alg:get_thermal_dists_hist}.

\begin{algorithm}[h]
\SetAlgoLined
\KwData{$T, D, I_{thermal}$}
\KwResult{$S_{comp}$}
Convert $T, D$ to NumPy arrays;\\
Initialize $S_{thermal}$ as zeroes($m, n$);\\
\For{each $t_i$ in $T$}{
  \For{each $d_j$ in $D$}{
    Extract $ROI_{t_i}, ROI_{d_j}$ from $I_{thermal}$;\\
    Calculate and normalize histograms $\hat{H}_{t_i}, \hat{H}_{d_j}$;\\
    Compute $s_{ij}$ using the Bhattacharyya coefficient and update $S_{thermal}[i, j]$;\\
  }
}
Compute $S_{comp} = \alpha \cdot S_{motion} + (1 - \alpha) \cdot S_{thermal}$;
\caption{Enhancing MOT in thermal imagery by leveraging motion and thermal similarities.}
\label{alg:get_thermal_dists_hist}
\end{algorithm}

\section{Results}
\label{sec:results}

\subsection{Benchmarking standard MOT models running on RGB}
The results of the standard implementations of ByteTrack and OCSORT on the RGB sequences of our newly collected dataset are given in Table~\ref{table_rgb_comparison}.
\begin{table*}[]
\centering

\begin{tabular}{|c|c|c|c|c|c|c|c|}
\hline
\textbf{Val RGB Sequence}       & \textbf{IDF1} & \textbf{IDP} & \textbf{IDR} & \textbf{Rcll} & \textbf{Prcn} & \textbf{MOTA} & \textbf{MOTP} \\ \hline
\textbf{2}  & 56.8\%        & 67.1\%       & 49.3\%       & 68.6\%        & 93.4\%        & 62.1\%        & 0.177         \\ \hline
\textbf{17} & 64.6\%        & 71.8\%       & 58.8\%       & 78.1\%        & 95.3\%        & 68.9\%        & 0.150         \\ \hline
\textbf{22} & 87.2\%        & 93.8\%       & 81.4\%       & 82.2\%        & 94.8\%        & 77.0\%        & 0.115         \\ \hline
\textbf{47} & 58.7\%        & 72.6\%       & 49.3\%       & 64.6\%        & 95.1\%        & 59.6\%        & 0.163         \\ \hline
\textbf{54} & 67.8\%        & 69.3\%       & 66.4\%       & 75.4\%        & 78.7\%        & 53.1\%        & 0.133         \\ \hline
\textbf{66} & 71.1\%        & 72.5\%       & 69.8\%       & 92.6\%        & 96.2\%        & 88.0\%        & 0.129         \\ \hline
\textbf{OVERALL}    & 63.6\%        & 72.5\%       & 56.7\%       & 72.4\%        & 92.7\%        & 64.7\%        & 0.152         \\ \hline
\end{tabular}

\bigskip

\begin{tabular}{|c|c|c|c|c|c|c|c|}
\hline
\textbf{Val RGB Sequence}       & \textbf{IDF1} & \textbf{IDP} & \textbf{IDR} & \textbf{Rcll} & \textbf{Prcn} & \textbf{MOTA} & \textbf{MOTP} \\ \hline
\textbf{2}  & 46.6\%        & 49.7\%       & 43.9\%       & 70.3\%        & 79.6\%        & 48.9\%        & 0.179         \\ \hline
\textbf{17} & 55.2\%        & 55.1\%       & 55.3\%       & 84.5\%        & 84.1\%        & 58.3\%        & 0.161         \\ \hline
\textbf{22} & 82.5\%        & 82.8\%       & 82.2\%       & 86.8\%        & 87.4\%        & 70.4\%        & 0.125         \\ \hline
\textbf{47} & 53.6\%        & 60.1\%       & 48.3\%       & 69.6\%        & 86.7\%        & 53.9\%        & 0.176         \\ \hline
\textbf{54} & 58.3\%        & 55.7\%       & 61.0\%       & 78.0\%        & 71.2\%        & 42.9\%        & 0.140         \\ \hline
\textbf{66} & 67.0\%        & 59.6\%       & 76.5\%       & 93.5\%        & 72.9\%        & 57.4\%        & 0.131         \\ \hline
\textbf{OVERALL}    & 56.8\%        & 58.6\%       & 55.1\%       & 76.3\%        & 81.0\%        & 53.7\%        & 0.160         \\ \hline
\end{tabular}
\caption{Results on the RGB validation sequences using the standard implementations of Bytetrack (top), and OCSORT (bottom).}

\label{table_rgb_comparison}
\end{table*}

\subsection{Weighted average alpha-value selection}
\begin{figure}[htbp]
    \centering
    \makebox[\linewidth][c]{%
        \includegraphics[width=0.6\linewidth]{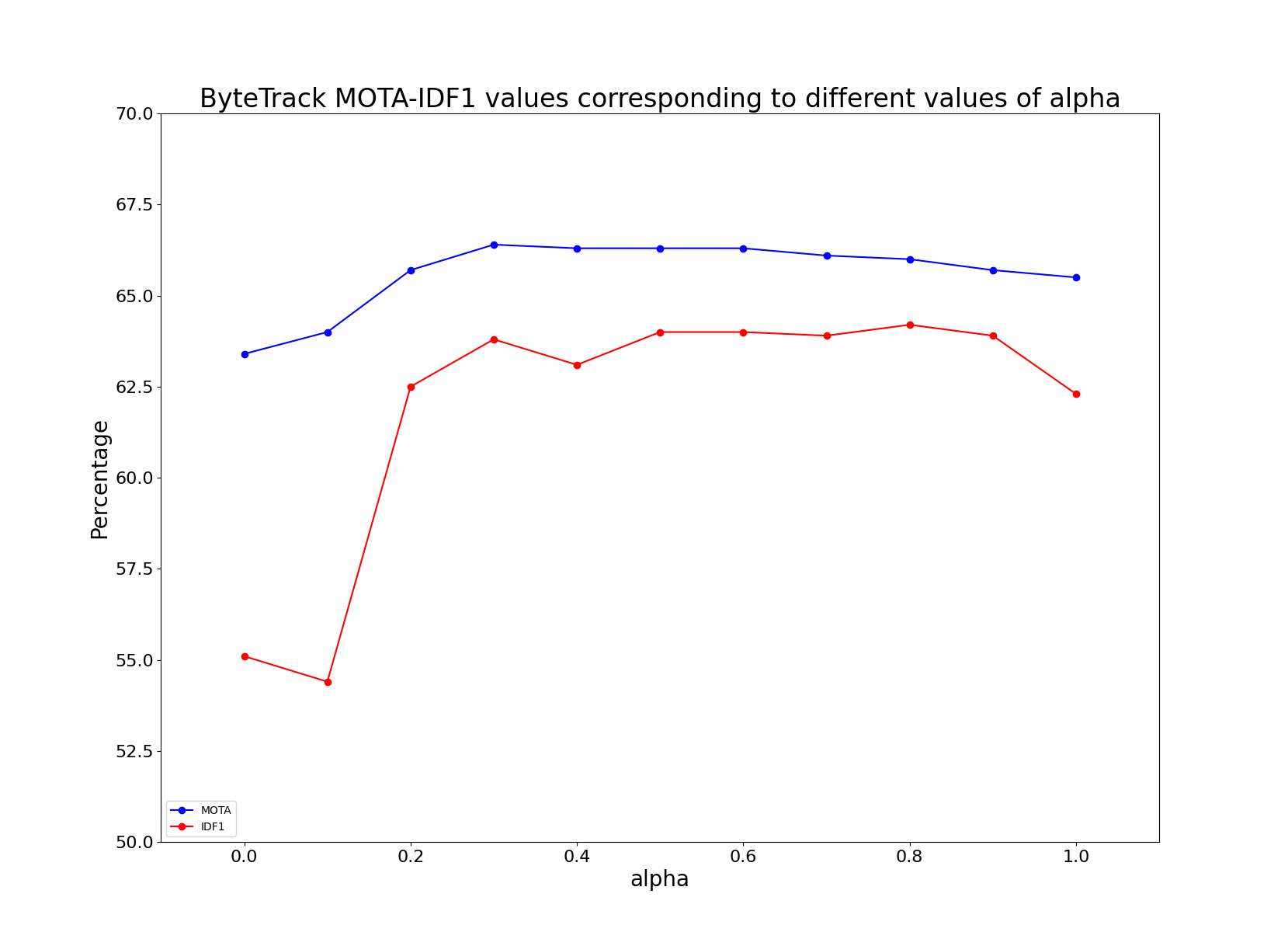}%
        \hfill
        \includegraphics[width=0.6\linewidth]{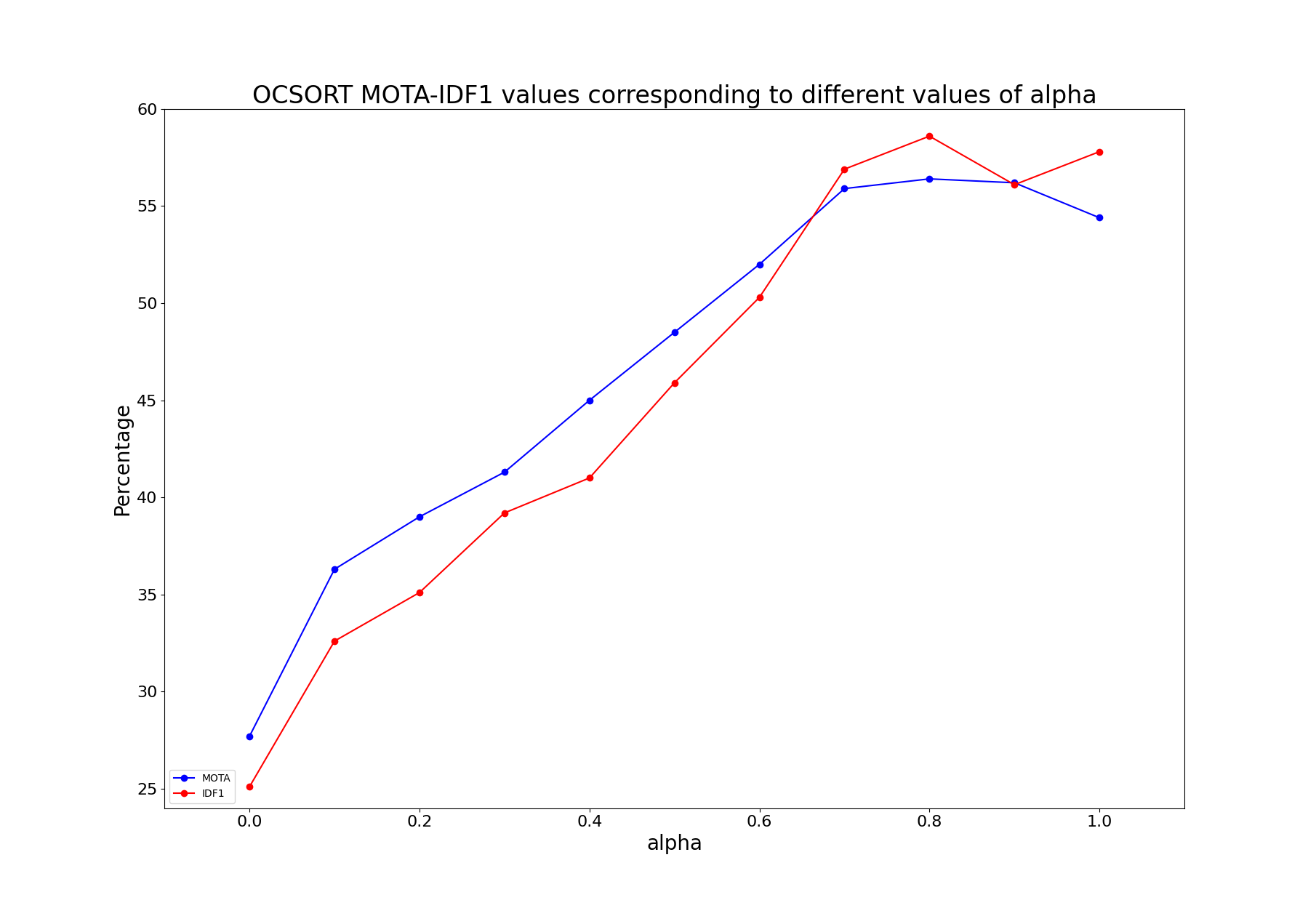}%
    }
    \caption{Left: MOTA and IDF1 values of our proposed box association method used with ByteTrack. Right: MOTA and IDF1 values used with OCSORT on the validation sequences of the RGB-Thermal MOT dataset using different values of alpha.}
    \label{figure_bytetrack_ocsort_alpha}
\end{figure}

In order to accurately find the best alpha-value (the weight of the motion distance matrix and the corresponding weight of the thermal distance matrix used to calculate the comprehensive distance matrix) we calculate the MOTA and IDF1 generated through our approach on the validation sequences of the RGB-Thermal MOT dataset. The results are given in figure ~\ref{figure_bytetrack_ocsort_alpha}.

The selected alpha-value for each model should ideally take into account the best trade-off between MOTA and IDF1. Analyzing figure \ref{figure_bytetrack_ocsort_alpha}, we can deduce the following:
\begin{itemize}
\item The overall performance of ByteTrack is proven to be better than OCSORT, with the maximum MOTA value achieved using ByteTrack reaching 66.4\%, while the maximum MOTA achieved using OCSORT is 56.4\%. 
\item A similar trend is observable through the analysis of the IDF1 values of both models, as the maximum IDF1 value achieved through ByteTrack is 64.2\% while the maximum IDF1 value achieved through OCSORT is 58.6\%.
\item For ByteTrack, the alpha value with the best trade-off between MOTA and IDF1 is 0.3, meaning the weight of motion association contribution to the comprehensive distance matrix is 30\%, while that of the thermal distance matrix is 70\%. This shows that ByteTrack benefits significantly from the thermal similarity matrix.
\item For OCSORT, the alpha value with the best trade-off between MOTA and IDF1 is 0.8, meaning the weight of motion association contribution to the comprehensive distance matrix is 80\%, while that of the thermal distance matrix is 20\%. This shows that OCSORT benefits significantly from the thermal similarity matrix.
\item For an alpha value of 0, meaning that only the thermal distance matrix is used for box association, ByteTrack achieves impressive results with 63.4\% MOTA and 55.1\% IDF1, surpassing the best performance of OCSORT. This shows the clear benefit of using thermal similarity for conducting box association, as it performs reasonably well even without utilizing motion association at all.
\item The difference in optimal alpha values between ByteTrack and OCSORT can be attributed to the distinct ways each algorithm integrates motion similarity and additional cues within their tracking frameworks. Despite both trackers emphasizing motion, the sensitivity of each to the incorporation of thermal similarity—via the alpha parameter—differs due to variations in their internal mechanisms and how they balance motion with other information sources. The observed discrepancy reflects the nuanced impact of thermal similarity on the trackers' performance, underscoring the necessity to tailor the alpha value to the specific architecture and processing strategy of each tracker to achieve optimal results.
\end{itemize}

\subsection{MOT Metrics Comparison}
\begin{figure}[htbp]
    \centering
    \includegraphics[width=1\linewidth]{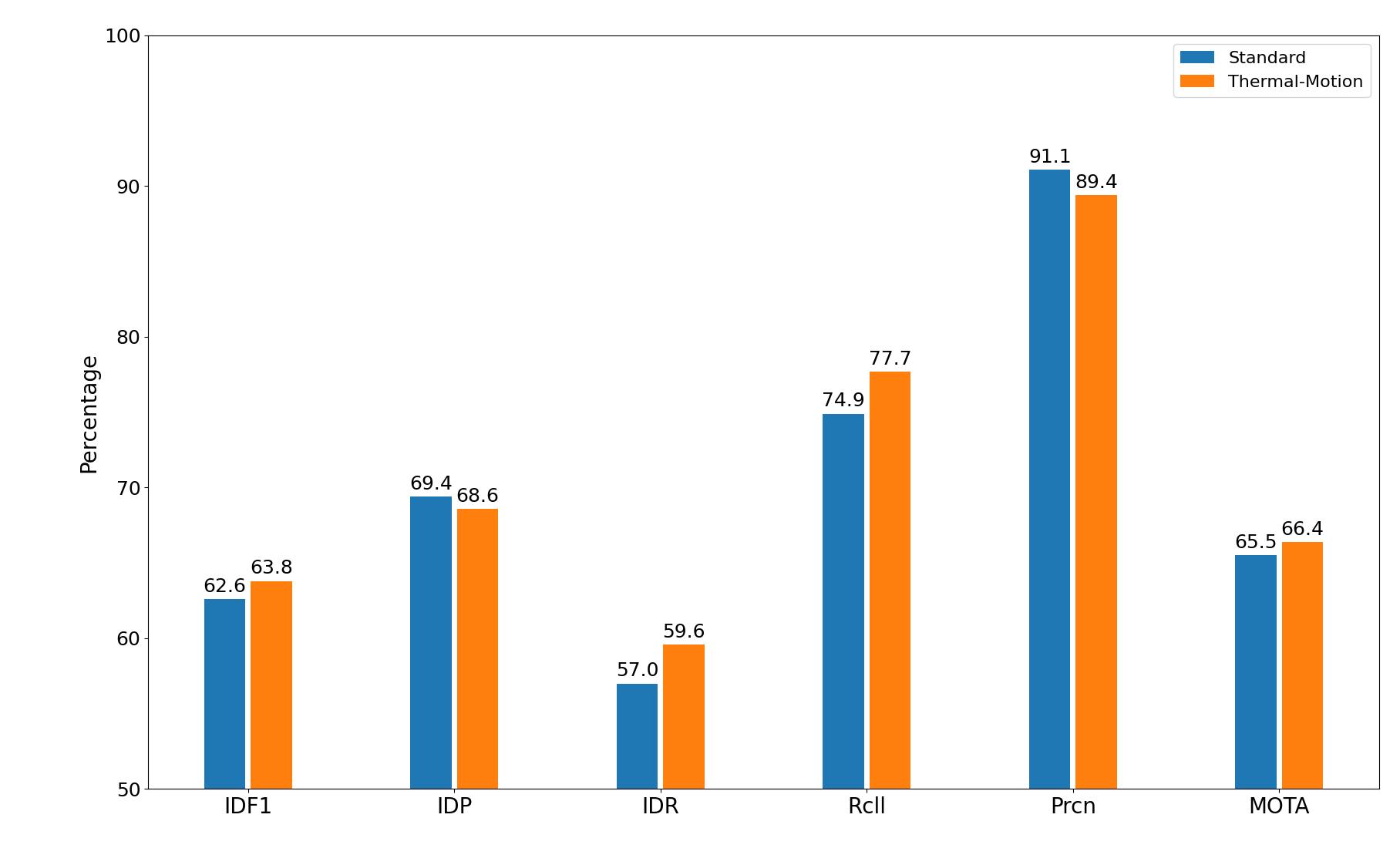}
    \caption{Overall comparison of the important MOT metrics running standard motion box association against our proposed box association algorithm using ByteTrack.}
    \vspace{2ex} % Adjust space as needed
    \includegraphics[width=1\linewidth]{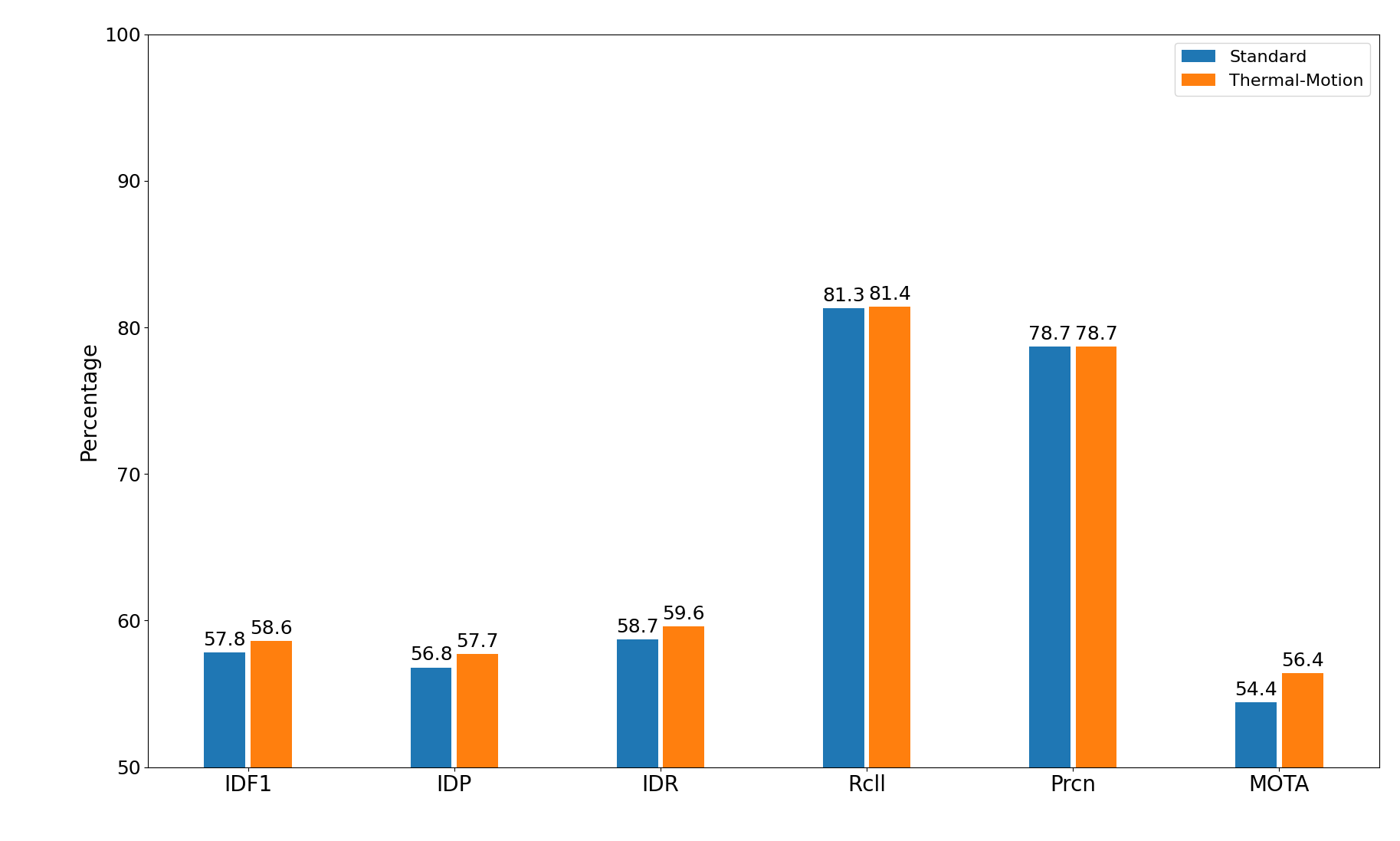}
    \caption{Overall comparison of the important MOT metrics running standard motion box association against our proposed box association algorithm using OCSort.}
    \label{figure_bytetrack_ocsort_graph}
\end{figure}

For evaluating the feasibility of our suggested box association method, we consider two state-of-the-art MOT models: Bytetrack~\cite{bytetrack} and OCSort~\cite{Cao2023}. Both Bytetrack and OCSort utilize motion association for the box-association step. The detailed results can be found in tables~\ref{table_bytetrack_comparison} and~\ref{table_ocsort_comparison}, while summarized metrics are given in figure \ref{figure_bytetrack_ocsort_graph}. \\

Analyzing the tables, the following conclusions can be made:
\begin{itemize}
\item ByteTrack benefits from utilizing thermal information when conducting box association. The overall MOTA and IDF1 for the original implementation of ByteTrack are 65.5\% and 62.6\% respectively. These increase to 66.4\% and 63.8\% respectively when using our proposed approach.
\item A similar trend can be seen with OCSORT, with the overall MOTA and IDF1 values increasing from 54.4\% and 57.8\% respectively in the original implementation of OCSORT to 56.4\% and 58.6\% respectively when using our proposed box association method. 
\end{itemize}

The results validate the effectiveness of our proposed model on enhancing tracking performance, where utilizing thermal similarity proves to be beneficial for box association when combined with motion similarity.

The proposed approach combines thermal and motion similarity scores through a weighted average. This fusion method effectively integrates two sources of information, allowing the system to make decisions based on both thermal and motion aspects.

By considering both thermal and motion aspects, the tracking system becomes more robust and adaptable. When thermal data indicates a strong match between objects with similar thermal signatures, the system can prioritize thermal information. Conversely, when motion cues are reliable, they can take precedence. This adaptability makes the tracking system more resistant to false positives and negatives. In addition, thermal data can assist in handling occlusions, a common challenge in MOT. When one object obscures another, thermal signatures may still be distinguishable, allowing the system to maintain the identity of both objects.

\begin{table*}[]
\centering

\begin{tabular}{|c|c|c|c|c|c|c|c|}
\hline
\textbf{Val Thermal Sequence}      & \textbf{IDF1} & \textbf{IDP} & \textbf{IDR} & \textbf{Rcll} & \textbf{Prcn} & \textbf{MOTA} & \textbf{MOTP} \\ \hline
\textbf{2}  & 51.2\%        & 69.3\%       & 40.6\%       & 56.7\%        & 96.7\%        & 52.5\%        & 0.184         \\ \hline
\textbf{17} & 71.5\%        & 79.1\%       & 65.2\%       & 75.8\%        & 92.0\%        & 65.7\%        & 0.173         \\ \hline
\textbf{22} & 75.4\%        & 84.9\%       & 67.9\%       & 74.4\%        & 93.0\%        & 67.3\%        & 0.163         \\ \hline
\textbf{47} & 59.1\%        & 62.2\%       & 56.4\%       & 81.5\%        & 89.9\%        & 70.1\%        & 0.170         \\ \hline
\textbf{54} & 63.8\%        & 65.7\%       & 62.0\%       & 77.8\%        & 82.4\%        & 59.2\%        & 0.176         \\ \hline
\textbf{66} & 67.9\%        & 75.5\%       & 61.7\%       & 78.5\%        & 95.9\%        & 74.4\%        & 0.154         \\ \hline
\textbf{OVERALL}        & 62.6\%        & 69.4\%       & 57.0\%       & 74.9\%        & 91.1\%        & 65.5\%        & 0.170         \\ \hline
\end{tabular}

\bigskip

\begin{tabular}{|c|c|c|c|c|c|c|c|}
\hline
\textbf{Val Thermal Sequence}      & \textbf{IDF1} & \textbf{IDP} & \textbf{IDR} & \textbf{Rcll} & \textbf{Prcn} & \textbf{MOTA} & \textbf{MOTP} \\ \hline
\textbf{2}  & 49.3\%        & 62.4\%       & 40.8\%       & 62.0\%        & 94.9\%        & 56.3\%        & 0.194         \\ \hline
\textbf{17} & 71.3\%        & 77.0\%       & 66.4\%       & 77.5\%        & 89.8\%        & 64.9\%        & 0.177         \\ \hline
\textbf{22} & 75.9\%        & 83.4\%       & 69.7\%       & 77.1\%        & 92.3\%        & 69.3\%        & 0.166         \\ \hline
\textbf{47} & 60.7\%        & 62.1\%       & 59.4\%       & 83.7\%        & 87.6\%        & 69.6\%        & 0.173         \\ \hline
\textbf{54}  & 71.4\%        & 71.5\%       & 71.3\%       & 81.2\%        & 81.4\%        & 60.9\%        & 0.183         \\ \hline
\textbf{66} & 68.0\%        & 74.3\%       & 62.6\%       & 80.2\%        & 95.1\%        & 75.3\%        & 0.157         \\ \hline
\textbf{OVERALL}        & 63.8\%        & 68.6\%       & 59.6\%       & 77.7\%        & 89.4\%        & 66.4\%        & 0.175         \\ \hline
\end{tabular}
\caption{Results of benchmarking Bytetrack on the thermal validation sequence using Bytetrack standard motion association (top), and using our novel thermal-motion box association method (bottom).}

\label{table_bytetrack_comparison}
\end{table*}

\begin{table*}[]
\centering

\begin{tabular}{|c|c|c|c|c|c|c|c|}
\hline
\textbf{Val Thermal Sequence}      & \textbf{IDF1} & \textbf{IDP} & \textbf{IDR} & \textbf{Rcll} & \textbf{Prcn} & \textbf{MOTA} & \textbf{MOTP} \\ \hline
\textbf{2}  & 51.7\%        & 60.4\%       & 45.2\%       & 66.3\%        & 88.8\%        & 51.3\%        & 0.197         \\ \hline
\textbf{17}   & 60.0\%        & 59.3\%       & 60.7\%       & 82.2\%        & 80.3\%        & 55.3\%        & 0.185         \\ \hline
\textbf{22} & 64.8\%        & 64.7\%       & 65.0\%       & 81.2\%        & 80.9\%        & 57.5\%        & 0.174         \\ \hline
\textbf{47} & 53.8\%        & 50.0\%       & 58.2\%       & 87.1\%        & 74.8\%        & 53.2\%        & 0.178         \\ \hline
\textbf{54} & 52.6\%        & 47.4\%       & 59.1\%       & 84.3\%        & 67.7\%        & 40.1\%        & 0.187         \\ \hline
\textbf{66} & 74.3\%        & 77.4\%       & 71.5\%       & 82.8\%        & 89.6\%        & 70.8\%        & 0.162         \\ \hline
\textbf{OVERALL}        & 57.8\%        & 56.8\%       & 58.7\%       & 81.3\%        & 78.7\%        & 54.4\%        & 0.180         \\ \hline
\end{tabular}

\bigskip

\begin{tabular}{|c|c|c|c|c|c|c|c|}
\hline
\textbf{Val Thermal Sequence}      & \textbf{IDF1} & \textbf{IDP} & \textbf{IDR} & \textbf{Rcll} & \textbf{Prcn} & \textbf{MOTA} & \textbf{MOTP} \\ \hline
\textbf{2}  & 42.4\%        & 49.6\%       & 37.1\%       & 66.4\%        & 88.9\%        & 54.1\%        & 0.198         \\ \hline
\textbf{17}    & 63.2\%        & 62.5\%       & 64.0\%       & 82.4\%        & 80.4\%        & 58.1\%        & 0.186         \\ \hline
\textbf{22} & 71.1\%        & 71.0\%       & 71.2\%       & 81.2\%        & 80.9\%        & 60.3\%        & 0.175         \\ \hline
\textbf{47} & 56.0\%        & 52.1\%       & 60.6\%       & 87.1\%        & 74.8\%        & 54.5\%        & 0.178         \\ \hline
\textbf{54} & 55.2\%        & 49.8\%       & 61.9\%       & 84.3\%        & 67.7\%        & 42.4\%        & 0.187         \\ \hline
\textbf{66} & 76.0\%        & 79.1\%       & 73.1\%       & 82.8\%        & 89.6\%        & 71. 7\%       & 0.162         \\ \hline
\textbf{OVERALL}        & 58.6\%        & 57.7\%       & 59.6\%       & 81.4\%        & 78.7\%        & 56.4\%        & 0.180         \\ \hline
\end{tabular}

\caption{Results of benchmarking OCSort on the thermal validation sequence using OCSort standard motion association (top), and using our novel thermal-motion box association method (bottom).}

\label{table_ocsort_comparison}
\end{table*}

\subsection{Limitations}
Our approach introduces an innovative use of thermal sensors for MOT, enhancing detection and tracking capabilities. However, it's pertinent to note the requirement of specialized thermal imaging equipment, which may not be universally accessible. Furthermore, the current application and validation of our methodology are confined to urban settings. The efficacy of our method in non-urban environments remains to be explored and would benefit from further diversification of the dataset to ensure broad applicability and robustness across varying scenarios.
\section{Conclusion}
\label{sec:conclusion}

In this paper, we have focused on enhancing the performance of MOT models operating in the thermal spectrum. Our key contribution lies in the introduction of a novel box association mechanism that harnesses both motion similarity and thermal object identity. This innovative approach enhances tracking accuracy and robustness by considering not just how objects move but also their distinct thermal signatures. The thermal and motion aspects are aggregated through a weighted average, resulting in a comprehensive similarity matrix that combines the strengths of both modalities. A key contribution of this work is that this novel box association method can be integrated with any two-stage MOT approach operating in the thermal spectrum, and encourages the exploration of utilization of unique spectrum characteristics when conducting box association. Given that two-stage MOT approaches are more robust and versatile than single-stage MOT models \cite{Zhang2020}, we believe this work could inspire more innovative research in this field.

In addition, we introduced the world's largest (to the best of our knowledge) dataset comprising of both RGB and corresponding thermal images, annotated for pedestrian MOT. We anticipate that this RGB-Thermal MOT dataset will be an invaluable resource for researchers in the fields of MOT and thermal vision perception. We fine-tuned state-of-the-art object detection models on this dataset, both for RGB and thermal images. Subsequently, we benchmarked leading MOT models on the dataset with and without our proposed box association method. 

The results are compelling. Notably, ByteTrack and OCSort, two state-of-the-art MOT models, exhibited improved performance when our proposed box association method was employed. The fusion of thermal and motion-based similarity scores proved advantageous, making the tracking system more adaptable, robust to occlusions, and resistant to false positives and negatives.

\bibliographystyle{splncs04}
\bibliography{main}
\end{document}